\documentclass{article}

\usepackage[preprint]{neurips_2026}

\usepackage[utf8]{inputenc} % allow utf-8 input
\usepackage[T1]{fontenc}    % use 8-bit T1 fonts
\usepackage{hyperref}       % hyperlinks
\usepackage{url}            % simple URL typesetting
\usepackage{booktabs}       % professional-quality tables
\usepackage{amsfonts}       % blackboard math symbols
\usepackage{nicefrac}       % compact symbols for 1/2, etc.
\usepackage{microtype}      % microtypography
\usepackage{xcolor}         % colors

\usepackage{multirow} % asra added  this
\usepackage{booktabs}
\usepackage{tabularx}
\usepackage{subcaption}
\usepackage{amsmath}
\usepackage{graphicx}
\usepackage{caption}

\usepackage{makecell}
\usepackage{xcolor}
\usepackage{threeparttable}

\title{Scalable Clinical Data Infrastructure and Comparative ML Evaluation for Hospitalisation Risk Prediction in Elderly Patients with Multiple Long-Term Conditions using CPRD}

\author{%
  \textbf{Asra Aslam$^1$},
  \textbf{Volodymyr Chapman$^2$},
  \textbf{Maurice M. O'Connell$^3$},
  \textbf{Aseel S. Abuzour$^4$}, \\
  \textbf{Michael Abaho$^5$},
  \textbf{Danushka Bollegala$^5$},
  \textbf{Gary Leeming$^5$},
  \textbf{Eduard Shantsila$^5$}, \\
  \textbf{Andrew Clegg$^4$},
  \textbf{Lauren E. Walker$^5$},
  \textbf{Iain Edward Buchan$^5$},
  \textbf{Samuel D. Relton$^2$} \\
  \vspace{0.5em} \\
  $^1$School of Information, University of Sheffield, Sheffield, United Kingdom \\
  $^2$School of Medicine, Faculty of Medicine and Health, University of Leeds, Leeds, United Kingdom \\
  $^3$Division of Informatics, University of Manchester, Manchester, United Kingdom \\
  $^4$Academic Unit for Ageing \& Stroke Research, University of Leeds, Leeds, United Kingdom \\
  $^5$Institute of Population Health, University of Liverpool, Liverpool, United Kingdom \\
  \vspace{0.5em} \\
  \texttt{a.aslam@sheffield.ac.uk}, 
  \texttt{V.Chapman@leeds.ac.uk}, 
  \texttt{maurice.oconnell@manchester.ac.uk}, \\
  \texttt{A.S.M.Abuzour@leeds.ac.uk}, 
  \texttt{micheal.abaho@liverpool.ac.uk}, 
  \texttt{danushka@liverpool.ac.uk}, \\
  \texttt{gary.leeming@liverpool.ac.uk}, 
  \texttt{Eduard.Shantsila@liverpool.ac.uk}, 
  \texttt{A.P.Clegg@leeds.ac.uk}, \\
  \texttt{md0u10pc@liverpool.ac.uk}, 
  \texttt{buchan@liverpool.ac.uk}, 
  \texttt{S.D.Relton@leeds.ac.uk}
}
\begin{document}

\maketitle

\begin{abstract}
Deep learning architectures are increasingly proposed for patient trajectory modeling in electronic health records (EHRs), yet their advantage over simpler, more interpretable models is rarely subjected to rigorous empirical scrutiny in real-world clinical settings. We present a comprehensive patient timeline pipeline applied to elderly patients in CPRD Aurum, incorporating 260 clinical conditions classified via a three-tier automated framework including specialised detection logic for 17 complex conditions. Using this infrastructure, we benchmark Temporal Graph Convolutional Neural Networks (TG-CNN) against Logistic Regression with LASSO regularisation and Random Forests for predicting 12-month all-cause emergency hospitalisation risk, motivated by (but not filtered to) the elevated risk of adverse drug reactions. Under cross-validation, TG-CNN achieves a marginally higher mean AUC-ROC than LASSO (0.712 vs. 0.705), whereas on the held-out test set LASSO achieves the highest discrimination of three models (AUC-ROC 0.733, versus 0.710 for Random Forest and 0.702 for TG-CNN). We show, that discrimination alone is an incomplete criterion for clinical deployment: after Platt calibration, LASSO is the only model with an acceptable calibration slope (0.817), while Random Forest (0.759) and, TG-CNN (0.391) remain substantially miscalibrated. We argue that LASSO, not the highest-discriminating model, is the model best suited to direct clinical deployment. We present lessons for the machine learning and healthcare community regarding data infrastructure, model selection, and value of calibration and interpretability in high-stakes decision support.
\end{abstract}

\section{Introduction}
Patients with multiple long-term conditions (MLTCs) face elevated treatment burden and heightened risk of harm through complex drug-drug and drug-disease interactions \citep{dumbreck2015interactions, osanlou2022adr}. Managing polypharmacy in this population is one of the most pressing challenges facing modern healthcare systems, with adverse drug reactions (ADRs) leading to hospitalisation estimated to cost the UK National Health Service \pounds2.21 billion per year \citep{chapman2026clustering}. Structured Medication Reviews (SMRs) offer a mechanism for medicines optimisation, but patient selection for these reviews remains subjective and inconsistent \citep{agwunobi2025smr, abuzour2024plosone}.

Electronic health records (EHRs) capture rich longitudinal patient data spanning diagnoses, prescriptions, laboratory tests, and clinical measurements. Yet systematic approaches to synthesising conditions, test results, and interventions over time remain limited. Health-AI Research \citep{walker2022dynairx} addresses this gap developing AI-based dynamic risk prediction and prescribing optimisation tools for patients with MLTCs, drawing on the Clinical Practice Research Datalink (CPRD), one of the world's largest primary care databases with records from over 60 million UK patients \citep{nhs_digital_ncras_cprd} with approximately 19 million in Aurum \citep{wolf2019cprd}. A central hypothesis motivating AI applications in this domain is that deep learning architectures capable of representing temporal sequences will outperform traditional statistical models, because the \emph{order} and \emph{timing} of clinical events carry prognostic information that regression-based approaches cannot fully exploit. Long Short-Term Memory (LSTM) networks \citep{hochreiter1997lstm} and graph neural network architectures have shown promise in benchmark tasks on EHR data. Temporal Graph Convolutional Neural Networks (TG-CNN) \citep{hancox2022tgcnn, aslam2025clustering} extend these ideas by explicitly modelling the elapsed time between events through a 3D tensor representation with exponential decay weighting, offering a theoretically compelling mechanism for learning clinically meaningful temporal patterns.

In this work we put this hypothesis to an empirical test. We apply TG-CNN to the CPRD elderly cohort, predicting all-cause hospitalisation within 12 months, and compare its performance against LASSO Logistic Regression \citep{tibshirani1996lasso} and Random Forests \citep{breiman2001rf}, two well-understood, interpretable baselines. We define the prediction target as 12-month all-cause emergency hospitalisation, excluding planned or elective admissions. This single outcome definition is used consistently across data labelling, model training, and evaluation. Elderly patients on complex polypharmacy are at elevated risk of hospitalisation due to adverse drug reactions, which motivates the clinical use case of structured medication review; however, adverse drug reactions are not used to filter or define the outcome label at any stage of the analysis. Our results reveal that the simpler models outperform TG-CNN on discrimination, and that the best-discriminating model fails on calibration. These investigations constitute clinically and scientifically important findings with direct implications for how the community should select and evaluate models for clinical deployment. The contributions of this research include:

%\subsection*{Contributions}
\begin{itemize}
  \item \textbf{Scalable CPRD infrastructure.} A reusable patient timeline pipeline for CPRD Aurum incorporating clinical conditions classified via a three-tier automated framework, with specialised detection logic for 17 complex conditions using validated clinical thresholds.

  \item \textbf{Large-scale benchmark.} A rigorous empirical comparison of temporal deep learning against interpretable baselines at a scale not previously reported in UK primary care EHR (570,125 training patients), with cross-validation for models.

  \item \textbf{Multi-criteria model evaluation framework.} Empirical evidence that AUC-ROC alone is an insufficient criterion for clinical model selection: we demonstrate that best-discriminating model requires post-hoc recalibration before deployment, while the most interpretable model does not, and propose a three-criteria evaluation standard.
\end{itemize}

\subsection*{Generalizable Insights about Machine Learning in the Context of Healthcare}
\begin{itemize}
  \item \textbf{Complexity does not guarantee superiority:} In real-world clinical datasets with sparse, irregular events and high dimensionality, interpretable models such as LASSO and Random Forests can match or exceed temporally complex deep learning architectures. The theoretical advantages of TG-CNN over simpler models do not automatically translate to empirical gains on routine EHR data.

  \item \textbf{Sparsity and data heterogeneity constrain temporal architectures:} Patient EHR timelines from primary care are characterised by highly sparse, sampled events and heterogeneous patient histories. These properties may neutralise the benefits of temporal weighting mechanisms that were designed for denser sequential data.

  \item \textbf{Calibration matters more than discrimination for clinical deployment:} A model that assigns systematically inflated risk scores cannot be used directly to communicate risk to clinicians without post-hoc correction. We show that the best-discriminating model in our comparison requires recalibration before deployment, while the most interpretable model does not. AUC-ROC alone should not determine model selection in clinical risk prediction.

%   % Change:
% We show that the best-discriminating model in our comparison requires recalibration
% before deployment, while the most interpretable model does not.

% % To:
% We show that both higher-complexity models in our comparison (Random Forest and
% TG-CNN) require recalibration before deployment, while the most interpretable
% model does not.
  \item \textbf{Interpretability has clinical value beyond performance:} In clinical decision support particularly for SMRs in primary care a model whose predictions can be explained to clinicians may be more deployable than marginally better-performing black box. Qualitative findings from research project (\citep{abuzour2026jmir}) place high value on understanding and trusting the basis of AI-generated recommendations.

  \item \textbf{Rigorous baseline comparison is essential:} Machine learning papers in health frequently compare against complex deep learning approaches. Our findings reinforce calls \citep{rudin2019stop, wynants2020prediction} for rigorous comparison against interpretable baselines before deploying complex AI in healthcare.
\end{itemize}
%rather than well-tuned classical models
%%%%%%%%%%%%%%%%%%%%%%%%%%%%%%%%%%%%%%%%%%%%%%%%%%%%%%%%%%%%%%%%%%%%%%%%%%%
%%%%%%%%%%%%%%%%%%%%%%%%%%%%%%%%%%%%%%%%%%%%%%%%%%%%%%%%%%%%%%%%%%%%%%%%%%%

\section{Related Work}
\subsection{Risk Prediction in Multimorbidity and Polypharmacy}

Existing risk prediction tools for populations with MLTCs predominantly target single endpoints. Tools such as QRISK and QAdmissions predict hospital admission or cardiovascular outcomes but do not model the dynamic, longitudinal interplay of conditions, medications, and outcomes over time. The consequence is that the rich temporal information embedded within EHRs remains largely underutilised for risk stratification \& treatment optimisation. A growing body of work has applied machine learning to EHR-based risk prediction. \citet{churpek2016ews} demonstrated that machine learning methods can outperform conventional regression for predicting clinical deterioration, but also showed that the margin over logistic regression was often modest. Studies of ADR risk and polypharmacy harm in older patients have relied on Cox regression and LASSO feature selection \citep{fahmi2023polypharmacy, chapman2026clustering}; these approaches have identified clinically meaningful patient subgroups associated with preventable ADR hospitalisations in the CPRD Aurum dataset. Qualitative research has elucidated both the appetite for AI tools in the SMR process and the substantial barriers to adoption \citep{abuzour2024plosone, abuzour2026jmir}. Healthcare professionals expressed scepticism about complex black-box predictive models and strongly preferred transparent/explainable systems. Visualising patient history timelines has also been studied to aid clinical reasoning in medication reviews \citep{hama2024visualization}.

%Visualisation of patient history timelines has also been studied as complementary approach to aiding clinical reasoning in medication reviews \citep{hama2024visualization}.

\subsection{Deep Learning for Patient Trajectories}

A substantial body of research has applied deep learning to EHR-based patient trajectory modelling. RETAIN \citep{choi2016retain}, a pioneering attention-based recurrent model, demonstrated interpretable prediction of heart failure onset from longitudinal EHR sequences by learning visit-level and code-level attention weights. Med2Vec \citep{choi2016med2vec} extended representation learning approaches to medical concepts and visits, showing that unsupervised embeddings of clinical codes could improve downstream prediction tasks. More recent transformer-based architectures such as BEHRT \citep{li2020behrt} and MedBERT \citep{rasmy2021med} adapted the BERT pre-training paradigm to EHR sequences, achieving strong performance on mortality and disease prediction benchmarks by jointly modelling clinical codes, temporal positions, and visit structure. 

%A substantial body of research has applied deep learning to EHR-based patient trajectory modelling. RETAIN \citep{choi2016retain}, a pioneering attention-based recurrent model, demonstrated interpretable prediction of heart failure onset from longitudinal EHR sequences by learning visit-level and code-level attention weights. Med2Vec \citep{choi2016med2vec} extended representation learning approaches to medical concepts and visits, showing that unsupervised embeddings of clinical codes could improve downstream prediction tasks. 

%More recent transformer-based architectures such as BEHRT \citep{li2020behrt} and MedBERT \citep{rasmy2021med} adapted the BERT pre-training paradigm to EHR sequences, achieving strong performance on mortality and disease prediction benchmarks by jointly modelling clinical codes, temporal positions, and visit structure.

However, many of these results were obtained on dense inpatient or secondary care datasets, such as MIMIC-III (ICU records), where events are frequent and temporally dense. Primary care EHRs have very different statistical properties: long observation windows spanning years to decades, irregular and infrequent visit patterns, and high dimensionality from hundreds of possible diagnosis and medication codes spread sparsely across patient histories. Evidence of deep learning superiority in this specific setting—large-scale, sparse, primary care, long-horizon—remains limited \citep{churpek2016ews, wynants2020prediction}. \citet{aslam2025clustering} adapted TG-CNN \citep{hancox2022tgcnn} for EHR trajectory clustering, demonstrating better-separated clusters than competing approaches on a proof-of-concept dataset and justifying its application to CPRD. However, cluster quality in an intermediate-complexity sequential dataset does not necessarily predict discrimination performance in a large-scale, sparse primary care EHR, especially for a binary prediction task such as all-cause emergency hospitalisation.

%However, many of these results were obtained on dense inpatient or secondary care datasets, such as MIMIC-III (ICU records), where events are frequent and temporally dense. Primary care EHRs have very different statistical properties: long observation windows spanning years to decades, irregular and infrequent visit patterns, and high dimensionality from hundreds of possible diagnosis and medication codes spread sparsely across patient histories. Evidence of deep learning superiority in this specific setting large-scale, sparse, primary care, long-horizon remains limited, with comparative studies finding that well-engineered classical models remain competitive \citep{churpek2016ews, wynants2020prediction}. 

%TG-CNN \citep{hancox2022tgcnn} was developed for online course drop-out prediction and subsequently adapted for EHR trajectory clustering by \citet{aslam2025clustering}. This work demonstrated that TG-CNN produces better-separated clusters than competing approaches on a proof-of-concept dataset, justifying its application to CPRD. However, cluster quality in an intermediate-complexity sequential dataset does not necessarily predict discrimination performance in a large-scale, sparse primary care EHR, especially for a binary prediction task such as all-cause emergency hospitalisation.

\subsection{Interpretable Versus Complex Models in Healthcare}
The problem between model complexity and clinical deployability has attracted increasing scholarly attention. \citet{rudin2019stop} argues compellingly that for high-stakes decisions in domains such as healthcare, interpretable models should be preferred unless complex models demonstrate a decisive performance advantage that cannot be achieved through careful feature engineering. \citet{wynants2020prediction} found that many published clinical prediction models suffer from methodological weaknesses, including failure to compare against well-tuned classical baselines. Taken together, these perspectives motivate a evaluation of when deep learning genuinely advances patient care versus when simpler approaches suffice.

A related and underappreciated problem is the over-reliance on discrimination metrics, particularly AUC-ROC, as the primary basis for model selection in clinical prediction. \citet{vancalster2019calibration} argue that a poorly calibrated model can cause systematic harm by misrepresenting absolute risk, and recommend calibration assessment as mandatory in clinical prediction model evaluation. \citet{steyerberg2010assessing} similarly demonstrate that calibration and clinical utility measures should complement discrimination in any rigorous evaluation framework. In practice, a model with AUC-ROC of 0.73 that systematically overestimates risk by 15 percentage points is less deployable than a model with AUC-ROC of 0.71 whose predicted probabilities are accurate, because clinicians and patients must act on absolute risk estimates, not on rankings alone. Our results provide an empirical case study of precisely this tension in a large-scale primary care EHR setting. The clustering work by \citet{chapman2026clustering}, conducted within the same CPRD cohort, used LASSO Cox regression to identify 74 features associated with ADR hospitalisation risk and subsequently derived five clinically interpretable patient subgroups. This research provides context: the predictive signal for adverse outcomes in this population can be captured effectively by relatively sparse feature sets extracted through penalised regression.

%%%%%%%%%%%%%%%%%%%%%%%%%%%%%%%%%%%%%%%%%%%%%%%%%%%%%%%%%%%%%%%%%%%%%%%%%%%
%%%%%%%%%%%%%%%%%%%%%%%%%%%%%%%%%%%%%%%%%%%%%%%%%%%%%%%%%%%%%%%%%%%%%%%%%%%

\section{Methodology}
%%%%%%%%%%%%%%%%%%%%%%%%%%%%%%%%%%%%%%%%%%%%%%%%%%%%%%%%%%%%%%%%%%%%%%%%%%%
\subsection{Overview of Clinical Codelist}

For preprocessing of CPRD data, we required comprehensive clinical codelists to identify patients with MLTCs. Traditional codelist development is extremely time-consuming, requiring months of manual expert review. We used the Generalised Codelist Automation Framework (GCAF) \citep{aslam2025automation} to address this challenge. GCAF combines automation with trusted sources (CALIBER and the Quality Outcomes Framework) to streamline codelist development. The framework automatically validates codes, removes duplicates from verified sources, and identifies new codes requiring clinical review. More than 90\% of codes were automatically verified using trusted sources and rest clincially validated.

%Both the DynAIRx codelists and the GCAF framework have been made publicly available \citep{aslam2025automation}.

%Using this approach, we generated approximately 214 condition-based codelists covering MLTCs, creating one of the most comprehensive MLTC codelists in the UK with approximately 14,000 SNOMED codes. Our automation framework reduced clinician validation time by over 80\%, completing final validation in only 7--9 hours of clinical expert time, compared to the months typically required by traditional methods. More than 90\% of codes were automatically verified using trusted sources before clinical validation. The remaining codes underwent systematic review by our expert team spanning mental health, primary care, pharmacy, and general practice. Both the DynAIRx codelists and the GCAF framework have been made publicly available \citep{aslam2025automation}.

%%%%%%%%%%%%%%%%%%%%%%%%%%%%%%%%%%%%%%%%%%%%%%%%%%%%%%%%%%%%%%%%%%%%%%%%%%%
\subsection{Data Preprocessing for CPRD}

We developed a comprehensive Patient Timeline pipeline to transform raw CPRD primary care records into structured timelines of medical events, conditions, and drug prescriptions. CPRD captures clinical events through several interconnected tables: patient identifiers, clinical observations (date, event, medcode, medcode-value), prescription records (prodcode, dosage, quantity, duration), and demographics including gender. These tables, stored in CSV or SQL format, provide the longitudinal clinical events, medications, and patient characteristics necessary for trajectory analysis. 

\subsubsection{Condition Categorisation Framework}
We classified medical conditions into three tiers based on diagnostic complexity: \textbf{Simple Conditions}: Identifiable directly from diagnostic codes in patient observation records via rule-based logic. A single diagnostic code was sufficient for classification. \textbf{Intermediate Conditions}: Required moderate clinical logic involving pattern checking in patient history or temporal relationships between events. Type 2 diabetes classification, for example, combined diagnostic codes with relevant laboratory test results recorded over time. \textbf{Complex Conditions}: Required sophisticated clinical logic functions integrating multiple data sources, applying clinical thresholds, and considering temporal patterns of measurements. We identified 17 conditions requiring complex detection logic: hypertension, hypotension, anaemia other, osteoporosis, cholesterol, chronic kidney disease, diabetes mellitus, falls, skin ulcer, smoking, alcohol problems, heart failure, obesity, chronic obstructive pulmonary disease (COPD), activity limitation, peripheral vascular disease, and thyroid problem.

\subsubsection{Complex Condition Detection Methodology}
For complex conditions, we implemented specialised clinical logic functions combining diagnostic codes with threshold-based clinical criteria. Each function implemented a two-stage strategy: first checking for explicit diagnostic codes, then assessing complex criteria such as aggregating measurements over time, applying clinical thresholds, or evaluating audit scores. Table~\ref{tab:complex_conditions} summarises the detection logic for all 17 complex conditions, providing a reusable reference for researchers applying this framework to other CPRD-based studies.

\begin{figure}
  \centering
  \includegraphics[width=0.95\textwidth]{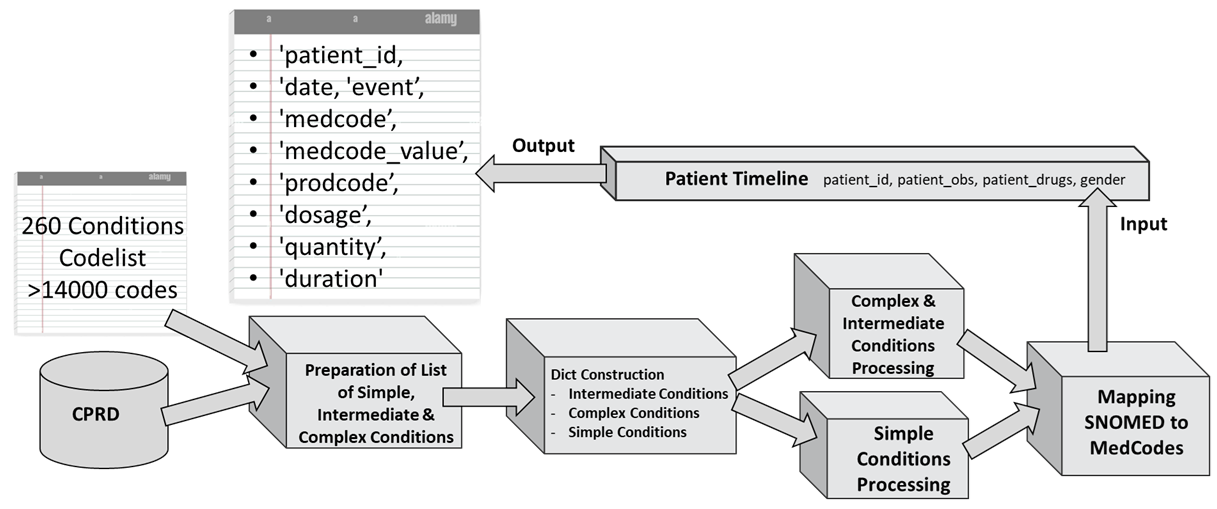}
  \caption{CPRD Data Preprocessing Workflow Architecture.}
  \label{fig:Preprocessing}
\end{figure}
% \begin{threeparttable}
% \end{threeparttable}
% \tnote{\textcolor{red}{*}}
% \begin{tablenotes}
%             %\small
%             \footnotesize
%             \item[\textcolor{red}{*}] TG-CNN requires temporal event sequences and cannot be trained on static demographic features alone. The Demographics-only baseline uses a neural network architecture identical to the demographic branch of the full TG-CNN model, providing a direct baseline of the architecture without a temporal signal.   
%         \end{tablenotes}
% Detection logic for 17 complex conditions (abbreviations defined below).
% Detection logic for the 17 complex conditions with described Abbreviations \tnote{\textcolor{red}{*}
%Detection logic for the 17 complex conditions (abbreviations here*)

\subsubsection{Processing Pipeline Structure}
The pipeline (shown in Fig.~\ref{fig:Preprocessing}) consisted of five main steps: (1) an input block retrieving raw CPRD data, (2) a mapping block standardising SNOMED to medcodes, (3) codelist preparation, (4) simple-conditions processing, and (5) complex and intermediate conditions processing, comprising dictionary construction mapping patient IDs to events and drugs, followed by execution of the tier-specific detection logic described above. The final output was a unified dataset with patient IDs, dates, events, medications, and condition labels annotated with both SNOMED and medcodes

\begin{table}[h]
\centering
\begin{threeparttable}
    \caption{Detection logic for the 17 complex conditions (abbreviations here\tnote{\textcolor{red}{*}}). }
    \label{tab:complex_conditions}
    \small
    \begin{tabularx}{\textwidth}{>{\raggedright\arraybackslash}p{3cm}|>{\raggedright\arraybackslash}p{3.5cm}|>{\raggedright\arraybackslash}X|>{\raggedright\arraybackslash}p{2.5cm}}
        \toprule
        \textbf{Condition} & \textbf{Primary Data Source} & \textbf{Threshold / Criterion} & \textbf{Guideline} \\
        \midrule
        Hypertension & ABPM / in-office BP & SBP $\geq$140 or DBP $\geq$90 mmHg ($\geq$3 readings) & NICE NG136 \\
        Hypotension & ABPM / cuff BP & SBP $\leq$90 or DBP $\leq$60 mmHg ($\geq$3 readings) & Clinical consensus \\
        Anaemia (Other) & Haemoglobin (Hb) & F: $<$11.5 g/dL; M: $<$13.0 g/dL & WHO \\
        Osteoporosis & BMD T-score (DEXA) & T-score $<$ $-$2.5 & WHO \\
        Cholesterol & HDL/LDL/Total/TG & Elevated in any component; gender-specific HDL & NICE CG181 \\
        CKD & eGFR & eGFR $<$60 mL/min/1.73m$^2$ (staged 3a--5) & KDIGO 2012 \\
        Diabetes Mellitus & HbA1c & HbA1c $\geq$6.5\% ($\geq$48 mmol/mol) & WHO/ADA \\
        Falls & Diagnostic codes + fall count & Any recorded fall incident & Clinical significance \\
        Skin Ulcer & Diagnostic codes + ulcer count & Any documented ulcer presence & Clinical significance \\
        Smoking & Tobacco codes (5-year window) & Any tobacco use within 5 years & Clinical consensus \\
        Alcohol Problems & AUDIT score / units/day & AUDIT $>$7 or hazardous intake & AUDIT (WHO) \\
        Heart Failure & LVEF (echocardiogram) & LVEF $<$40\% (HFrEF) & ESC 2021 \\
        Obesity & BMI & BMI $\geq$30 kg/m$^2$ & WHO \\
        COPD & Diagnostic codes + exacerbations & $\geq$1 recorded acute exacerbation & GOLD \\
        Activity Limitation & Barthel Index & Score $\leq$18 (out of 20) & Barthel 1965 \\
        Peripheral Vascular Disease & ABPI & ABPI $<$0.95 & AHA/ACC \\
        Thyroid Problem & TSH & TSH $<$0.36 or $>$5.5 mIU/L & Local lab reference \\
        \bottomrule
    \end{tabularx}
    \begin{tablenotes}
        %\small
        \footnotesize
        %\tiny
        \item[\textcolor{red}{*}] SBP: systolic blood pressure; DBP: diastolic blood pressure; ABPM: ambulatory blood pressure monitoring; DEXA: dual-energy X-ray absorptiometry; eGFR: estimated glomerular filtration rate; LVEF: left ventricular ejection fraction; ABPI: ankle-brachial pressure index; TSH: thyroid-stimulating hormone; F: female; M: male.   
    \end{tablenotes}
\end{threeparttable}
\end{table}

%The pipeline (shown in Fig.~\ref{fig:Preprocessing}) consisted of five main steps: (1) Input block retrieving raw CPRD data, (2) Mapping block standardising SNOMED to medcodes, (3) Codelist preparation creating the 260-condition reference list, (4) Simple conditions processing filtering records matching simple condition codes, and (5) Complex and intermediate conditions processing comprising dictionary construction mapping patient IDs to events and drugs, intermediate conditions processing, and complex conditions executing the 17 specialised clinical logic functions. The final output was a unified dataset with patient IDs, dates, events, medications, and condition labels annotated with both SNOMED and medcodes.

%%%%%%%%%%%%%%%%%%%%%%%%%%%%%%%%%%%%%%%%%%%%%%%%%%%%%%%%%%%%%%%%%%%%%%%%%%%
\subsection{Optimisation for Training}

Prior to training, we performed extensive optimisation of the training data to ensure efficient processing and effective temporal representation. This multi-stage transformation converted raw CPRD patient data into a structured format suitable for graph-based temporal learning.

\noindent\textbf{Data Loading and Filtering:} We implemented a preprocessing pipeline operating in both training and test modes, loading patient IDs accordingly for each phase. Raw event data and label data were read from parquet files using Dask for efficient handling of large-scale datasets. The data were filtered to retain only necessary columns: \texttt{patid} (patient ID), \texttt{event} (clinical event code), \texttt{eventdate} (event date), and \texttt{label} (outcome label). Event data were merged with corresponding label data to create a unified patient-event dataset. All date fields were converted to timestamps, and patient and event identifiers were encoded as integer values to facilitate matrix-based computations.

\noindent\textbf{Temporal Representation via Timestamp Creation:} A critical step involved converting event dates into integer timestamps for consistent temporal analysis. We defined a reference start date of 1 January 1955 and calculated the number of days elapsed between each event date and this reference point. Missing or invalid timestamps were filled with zero and cast as integers, establishing a consistent temporal scale across all patient trajectories. The temporal weighting applied to each edge of the patient event graph is given by $G(i,j,k) = \exp(-\gamma t)$, where $t$ denotes the temporal interval between events, and $\gamma$ is a learnable parameter optimized jointly with the network during training rather than a fixed decay constant. This learnable-decay formulation mechanism is detailed in the methodology proposed by \citet{hancox2024}.

%For example, 27 September 1957 corresponded to timestamp 1000, representing 1,000 days since the reference. 

% \noindent\textbf{Trajectory Index and Value Construction:} For each patient, we generated parallel data structures capturing both event sequence and temporal spacing. Processing was performed in memory-efficient chunks of exactly 25,000 patients at a time. An exponential decay transformation $\exp(-t/1000)$ was applied to timestamps, assigning higher weights to more recent events whilst maintaining sensitivity to temporal distances. The trajectory representation comprised two parallel structures: an \emph{indices} list of triplets $[\text{patient\_id}, \text{event\_id}, \text{timestep}]$ where \texttt{timestep} denotes the 1-based sequential position of the event, and a \emph{values} list of corresponding exponentially decayed timestamps. For example, a patient's first event (event ID 5) and second event (event ID 7) would produce indices \texttt{[[0,5,1],[0,7,2]]} and values \texttt{[0.9, 0.8]}, where 0.9 and 0.8 represent the decayed timestamps respectively. This dual structure enables TG-CNN to simultaneously process both the sequential ordering and temporal spacing of clinical events.

% This mechanism is documented in the methodology paper by Hancox et al. (2024, IEEE; https://ieeexplore.ieee.org/abstract/document/10903462), specifically in Methods Section E. The camera-ready paper will include this description of learnable parameterisation in Section 3.3 and cite Hancox et al. 2024.

\noindent\textbf{Dataset Structuring:} A comprehensive preprocessed DataFrame was constructed integrating trajectory indices with temporal values. For each patient, the dataset contained four fields: (1) a \texttt{user} column with the patient ID, (2) an \texttt{indices} column storing the list of event-timestep triplets, (3) a \texttt{values} column containing the exponentially decayed timestamp values, and (4) a \texttt{num\_timesteps} column recording the total number of events in that patient's trajectory. This structure provided the TG-CNN with complete information about event sequences, temporal patterns, and trajectory lengths per patient.

\noindent\textbf{Label Assignment:} For supervised learning, a single outcome label was assigned to each patient based on their most recent timestamped event, identified using \texttt{numpy}'s \texttt{argmax} function over timestamps. Each patient is assigned a fixed index date of 1 April 2019. Predictor features are derived from all clinical records up to and including 31 March 2019. The outcome label is defined as an all-cause emergency hospital admission within the twelve-month outcome window immediately following the index date (1 April 2019 to 31 March 2020). No event occurring on or after 1 April 2019 is used to construct predictor features. This ensured that the prediction target reflected the patient's most current clinical state, most relevant for prospective risk prediction. The complete preprocessed dataset containing patient identifiers, event sequences with temporal information, and outcome labels was saved to disk for efficient loading during model training.

%For supervised learning, a single outcome label was assigned to each patient based on their most recent timestamped event, identified using \texttt{numpy}'s \texttt{argmax} function over timestamps. Each patient is assigned a fixed index date of 1 April 2019, with predictor and outcome windows defined as disjoint per Section \ref{sec:CohortSelection}. This ensured that the prediction target reflected the patient's most current clinical state, most relevant for prospective risk prediction. The complete preprocessed dataset containing patient identifiers, event sequences with temporal information, and outcome labels was saved to disk for efficient loading during model training.

%Each patient is assigned a fixed index date of 1 April 2019. Predictor features are derived from all clinical records up to and including 31 March 2019. The outcome label is defined as an all-cause emergency hospital admission within the twelve-month outcome window immediately following the index date (1 April 2019 to 31 March 2020). No event occurring on or after 1 April 2019 is used to construct predictor features. 

%The outcome label is defined as an unplanned hospital admission for an adverse drug reaction within the twelve-month outcome window immediately following the index date (1 April 2019 to 31 March 2020). 

%%%%%%%%%%%%%%%%%%%%%%%%%%%%%%%%%%%%%%%%%%%%%%%%%%%%%%%%%%%%%%%%%%%%%%%%%%%
\subsection{Temporal Graph Convolutional Neural Network (TG-CNN)}
\label{sec:tgcnn}
TG-CNN \citep{hancox2022tgcnn, aslam2025clustering} models patient trajectories as temporal graphs in which nodes represent clinical events and edges encode both sequential order and elapsed time. The architecture introduces a three-dimensional tensor representation that retains the sequential order of clinical actions alongside the elapsed time between them. An exponential decay function applied to temporal intervals assigns differential importance to events based on their timing, enabling the model to distinguish short-term acute changes from long-term disease progression patterns.

%\subsubsection{TG-CNN Architecture and Training}

\noindent\textbf{Model Architecture.} The TG-CNN architecture (in Appendix \ref{appendix_A} \ref{fig:config_tgcnn}) integrated several components designed to capture both spatial and temporal patterns in patient trajectories. Long Short-Term Memory (LSTM) layers modelled sequential dependencies across clinical events, enabling the network to learn how event sequences evolved over time. Gaussian Error Linear Unit (GELU) activation functions provided smooth non-linearities and improved gradient propagation compared to ReLU activations. The model explicitly incorporated temporal information through the exponential decay weighting of timestamps described above, assigning greater importance to recent events whilst maintaining temporal aspects. 
%traditional 
%Two configurations were evaluated: a single TG-CNN layer and a dual-layer two-stream architecture (Figure~\ref{fig:config_tgcnn}), with the latter allowing parallel processing of different temporal feature representations.

\noindent\textbf{Stratified Sampling and Batch Construction.} Training data were loaded from the preprocessed parquet files using distributed computing frameworks. The structured feature representations, stored as index-value pairs, were deserialised and converted into 3D sparse tensors. Given the inherent sparsity of clinical data where individual patients experience only a small fraction of all possible medical events the sparse tensor representation efficiently encoded patient trajectories whilst minimising memory overhead. Class distribution analysis confirmed imbalanced outcome prevalence (a common characteristic of clinical datasets), so we employed stratified sampling when partitioning data into training and validation sets, maintaining proportional outcome class representation in both. Within each batch, class frequencies were calculated and inverse-frequency weighting applied to loss contributions, amplifying the influence of underrepresented outcome classes and preventing the model from converging to trivial majority-class predictions.

\noindent\textbf{Regularisation and Hyperparameters.} To prevent overfitting, multiple regularisation strategies were applied: dropout randomly deactivated neurons during training, and both L1 and L2 weight regularisation penalised large parameter values to encourage simpler, more generalisable representations. Hyperparameters included: learning rate (gradient descent step sizes), dropout probability (controlling regularisation strength), LSTM hidden dimensions (determining sequential memory capacity), and L1/L2 regularisation coefficients. These were selected through preliminary experiments and refined on held-out validation set.

\noindent\textbf{Training and Validation Procedure.} During each training epoch, a custom training step function coordinated forward propagation through the sparse tensors, loss computation, backpropagation, and parameter update via gradient descent. During forward propagation, intermediate feature representations and their corresponding ground-truth labels were extracted and stored from internal layers, enabling post-hoc interpretability analysis of learned representations at different abstraction levels. After each epoch, performance was assessed on the validation set with model parameters fixed, producing training and validation metric curves. If validation performance failed to improve past a patience threshold across consecutive epochs, training was halted via early stopping to prevent overfitting.

%During each training epoch, a custom training step function coordinated forward propagation through the sparse tensors, loss computation, backpropagation, and parameter update via gradient descent. During forward propagation, both intermediate feature representations and their corresponding ground-truth labels were extracted and stored from internal network layers, enabling post-hoc interpretability analysis of learned representations at different abstraction levels. After each epoch, performance was assessed on the validation set with model parameters fixed, producing training and validation metric curves. If validation performance failed to improve beyond a predefined patience threshold across consecutive epochs, training was halted via early stopping to prevent overfitting.

%\textbf{Performance Metrics.} We monitored accuracy, loss, precision, recall, F1 score, and AUC-ROC throughout training, computed both overall and per class. All metrics were logged at each epoch, creating a longitudinal record of learning dynamics. The model state achieving optimal validation AUC-ROC was saved and used for all downstream evaluation.

% \begin{figure}[!ht]
%   \centering
%   \includegraphics[width=0.99\textwidth]{images/Configuration 1 Single TGCNN Layer.png}
%   \caption{Configuration 1: Single TG-CNN Layer Architecture.}
%   \label{fig:config_1}
% \end{figure}

%%%%%%%%%%%%%%%%%%%%%%%%%%%%%%%%%%%%%%%%%%%%%%%%%%%%%%%%%%%%%%%%%%%%%%%%%%%

\subsection{Baseline Models}

%\subsubsection{Logistic Regression with LASSO Regularisation}

\noindent\textbf{Logistic Regression with LASSO Regularisation}: We trained LASSO Logistic Regression \citep{tibshirani1996lasso} using scikit-learn \citep{pedregosa2011sklearn}. Features included patient demographics and binary/count representations of clinical events extracted from the preprocessed CPRD timelines. Optimal regularisation strength was selected via 5-fold cross-validation. The LASSO penalty performs simultaneous feature selection and coefficient shrinkage, producing a sparse model with non-zero coefficients only for the most predictive features a particularly desirable property in clinical deployments where model transparency is valued by end-users.

%\subsubsection{Random Forest}
\noindent\textbf{Random Forest}: We trained a Random Forest classifier \citep{breiman2001rf} on the same feature set. Random Forests provide non-linear decision boundaries and implicit feature importance estimates without the stringent linearity assumptions of logistic regression. Hyperparameters (number of trees, maximum depth) were tuned via 5-fold cross-validation.

%%%%%%%%%%%%%%%%%%%%%%%%%%%%%%%%%%%%%%%%%%%%%%%%%%%%%%%%%%%%%%%%%%%%%%%%%%%
%%%%%%%%%%%%%%%%%%%%%%%%%%%%%%%%%%%%%%%%%%%%%%%%%%%%%%%%%%%%%%%%%%%%%%%%%%%

\section{Cohort}

\subsection{Cohort Selection}
\label{sec:CohortSelection}
Our cohort was drawn from the CPRD Aurum dataset \citep{wolf2019cprd}, a UK primary care database of anonymised electronic health records. We focused on an elderly cohort, defined as patients aged 65 years or older at the start of the observation period, consistent with the clinical focus on older patients with MLTCs and frailty. Predictor features (comprising condition codes, medication codes, and demographic characteristics) are derived exclusively from clinical records dated on or before 31 March 2019. The outcome observation window spanned 1 April 2019 to 31 March 2020, with the primary outcome defined as an all-cause emergency hospital admission, identified via NHS admission method codes 21–25, 28, and 2A–2D, excluding planned or elective admissions. These predictor and outcome windows were completely disjoint by design, ensuring that no clinical event occurring on or after 1 April 2019 was included in the predictor feature set. This period was chosen for data completeness and to avoid confounding from the COVID-19 pandemic.

%Our cohort was drawn from the CPRD Aurum dataset \citep{wolf2019cprd}, a UK primary care database of anonymised electronic health records. We focused on an elderly cohort, defined as patients aged 65 years or older at the start of the observation period, consistent with the clinical focus on older patients with MLTCs and frailty. Predictor features (comprising condition codes, medication codes, and demographic characteristics) are derived exclusively from clinical records dated on or before 31 March 2019. Each patient was assigned a fixed index date of 1 April 2019. The twelve-month outcome window spanned 1 April 2019 to 31 March 2020, with the primary outcome defined as an all-cause emergency hospital admission, identified via NHS admission method codes 21–25, 28, and 2A–2D, excluding planned or elective admissions. These predictor and outcome windows were completely disjoint by design, ensuring that no clinical event occurring on or after 1 April 2019 was included in the predictor feature set. This period was chosen for data completeness and to avoid confounding from the COVID-19 pandemic.

%The outcome observation window spanned 1 April 2019 to 31 March 2020, with the primary outcome defined as an all-cause emergency hospital admission, identified via NHS admission method codes 21–25, 28, and 2A–2D, excluding planned or elective admissions.

\subsection{Data Extraction}

Raw data were extracted from CPRD Aurum tables covering patient demographics, clinical observations, drug prescriptions, and linked secondary care (Hospital Episode Statistics) data for hospitalisation outcomes. Patient conditions were identified using codelists \citep{aslam2025automation}. Prescriptions were linked to the British National Formulary (BNF) hierarchy to create medication feature categories.

\subsection{Feature Choices}

For the interpretable baseline models, features included: patient demographics (age, gender, Index of Multiple Deprivation), binary indicators for each of the clinical conditions (derived via the preprocessing pipeline), binary indicators for BNF medication categories, and interaction features combining high-risk conditions with specific medication classes known to be associated with adverse outcomes (more details in Appendix Table~\ref{tab:feature_inventory}). For TG-CNN, the temporal event sequence and timestamps were provided as the primary input, with demographics incorporated as static features. Training data comprised $N = 570{,}125$ patients and test data comprised $N = 63{,}347$ patients, split with stratification on hospitalisation outcome to maintain proportional representation.

\subsection{Cohort and Pipeline Characterisation}
We characterise the cohort and data properties with summary shown in Table \ref{tab:cohort}, which contextualise the modelling results and will support reproducibility for researchers applying this infrastructure to other CPRD-based studies. The sparsity of patient timelines deserves particular emphasis. CPRD primary care records are characterised by long observation windows spanning years to decades, with clinical events recorded only at point of contact. The majority of patients have events distributed across hundreds of possible clinical codes, but any individual patient activates only a small fraction of the full event space. This sparsity is a structural property of primary care EHR data that distinguishes it sharply from the dense, regularly sampled sequences for which most temporal deep learning architectures were designed, with the results reported in Section~\ref{sec:quantresults}.
%We characterise the cohort and data properties produced by the pipeline ()

%%%%%%%%%%%%%%%%%%%%%%%%%%%%%%%%%%%%%%%%%%%%%%%%%%%%%%%%%%%%%%%%%%%%%%%%%%%
%%%%%%%%%%%%%%%%%%%%%%%%%%%%%%%%%%%%%%%%%%%%%%%%%%%%%%%%%%%%%%%%%%%%%%%%%%%

\begin{table}
\centering
    \caption{Cohort and pipeline summary. Exact class distribution is reflected in the stratified sampling strategy described in Section~\ref{sec:tgcnn}.}
    \label{tab:cohort}
    \begin{tabularx}{\textwidth}{l X}
        \toprule
        \textbf{Characteristic} & \textbf{Value} \\
        \midrule
        Total patients (training set) & 570,125 \\
        Total patients (test set) & 63,347 \\
        Age criterion & $\geq$65 years at 01 April 2019 \\
        Observation window & 01 April 2019 to 31 March 2020 \\
        Outcome: all-cause hospitalisation (training) & $\sim$20\% (class imbalance present) \\
        Clinical conditions modelled & Simple, Intermediate, Complex tiers) \\
        Complex conditions with bespoke logic & 17 \\
        Feature space: LASSO / RF & Demographics + Condition indicators + BNF medication categories \\
        Feature space: TG-CNN & Temporal event sequence + timestamps + static demographics \\
        COVID-19 confounding & Avoided by restricting window to pre-March 2020 \\
        \bottomrule
    \end{tabularx}
        
\end{table}

\section{Results}

%\subsection{Training Optimisation and Convergence Analysis}
\subsection{Quantitative Results}
\label{sec:quantresults}
%Before comparing model performance, it is worth contextualising the effort invested in optimising TG-CNN for this task. 
The CPRD elderly cohort presents a challenging computational setting: the training set of $N = 570{,}125$ patients, each with event trajectories spanning multiple years and hundreds of possible clinical event types, required careful data engineering to fit within GPU memory. Chunked processing of 25,000 patients at a time and sparse tensor representation were both necessary to make training tractable. For the TG-CNN, we explored different configurations  and conducted hyperparameter sweeps over learning rate, LSTM hidden dimensions, dropout probability, and regularisation coefficients. Training was monitored via validation AUC-ROC with early stopping applied. Convergence was slow relative to the baseline models, reflecting the increased parameter count and the sparsity of the input data (Appendix \ref{appendix_A} Fig \ref{fig:config_tgcnn} and \ref{fig:all_tgcnn_combined}). For the baseline models, LASSO Logistic Regression and Random Forest were both tuned similarly. The LASSO regularisation strength was selected over a logarithmic grid; Random Forest hyperparameters (number of trees: 500, maximum features per split: $\sqrt{p}$) were chosen to balance variance reduction with computational cost. Both models converged reliably and exhibited low cross-fold variance (standard deviations $\leq$0.002), indicating stable performance estimates. All models were evaluated using 5-fold stratified cross-validation, with final performance assessed on the held-out test set. The primary metric was the AUC-ROC (area under the receiver operating characteristic curve), which measures overall discrimination ability across classification thresholds and is robust to class imbalance. We also report standard deviation across folds to characterise stability.

\begin{table} %[htbp]
\centering
\begin{threeparttable}
  \caption{Model performance summary: mean validation AUC-ROC and standard deviation under 5-fold cross-validation. All models predict all-cause hospitalisation within 12 months in an elderly primary care cohort ($N_{\text{train}} = 570{,}125$).}
  \label{tab:model_performance}

  \begin{tabularx}{\textwidth}{l l c | c}
  \toprule
  & & \multicolumn{2}{c}{\textbf{5-fold Cross-Validation}}\\
  
  \textbf{Model} & \textbf{Data} & \textbf{Mean } & \textbf{Standard} \\
  & & \textbf{AUC-ROC} & \textbf{Deviation} \\

  \midrule

  \multirow{3}{*}{{Logistic Regression}}
   & Demographics & 0.640 & 0.0013 \\
   \cmidrule(lr){2-4}
   \multirow{2}{*}{(LASSO)}& Events & 0.688 & 0.0056 \\
   \cmidrule(lr){2-4}
  & Demographics + events & 0.705 & 0.0041 \\

  \midrule

  \multirow{3}{*}{Random Forest}
   & Demographics & 0.643 & 0.0024 \\
   \cmidrule(lr){2-4}
   & Events & 0.721 & 0.0014 \\
   \cmidrule(lr){2-4}
   & Demographics + events & 0.734 & 0.0010 \\

  \midrule

  \multirow{3}{*}{TG-CNN}
   & Demographics\tnote{\textcolor{red}{*}} (NN baseline,& \multirow{2}{*}{0.658} & \multirow{2}{*}{0.0040} \\
   &TG-CNN Not Applicable)& & \\
   \cmidrule(lr){2-4}
   & Events & 0.685 & 0.0029 \\
   \cmidrule(lr){2-4}
   & Demographics + events & 0.712 & 0.0022 \\

  \bottomrule
  \end{tabularx}

  \begin{tablenotes}
    %\small
    \footnotesize
    \item[\textcolor{red}{*}] TG-CNN requires temporal event sequences and cannot be trained on static demographic features alone. The Demographics-only baseline uses a neural network architecture identical to the demographic branch of the full TG-CNN model, providing a direct baseline of the architecture without a temporal signal.
     
  \end{tablenotes}
\end{threeparttable}
\end{table}
%\scriptsize
%\tiny
%**TG-CNN results are from cross-validated runs.

%\item[\textcolor{red}{**}] TG-CNN results are evaluated under the same 5-fold cross-validation protocol as the LASSO and Random Forest baselines.

Table \ref{tab:model_performance} presents the models comparison with results across all feature configurations. LASSO Logistic Regression trained on event features alone achieved a mean AUC-ROC of 0.688 ($\sigma =  0.0056$), while TG-CNN trained on events reaches till 0.685. On the combined demographics and events feature set, LASSO achieved 0.705 ($\sigma = 0.0041$), while Random Forest achieved the highest overall AUC-ROC of 0.734 ($\sigma =  0.0010$). Because TG-CNN requires temporal event sequences and cannot be applied directly to static-only inputs, static demographic performance is reported using a static neural network baseline (Table~\ref{tab:model_performance}). This baseline is structurally identical to the demographic branch of the fused TG-CNN architecture, providing a direct evaluation of the model's static feature processing without a temporal signal. TG-CNN on the combined feature set achieved 0.712 ($\sigma = 0.0022$) while lagging behind Random Forest (0.734). For the LASSO model, 548 non-zero coefficients were retained after regularisation, indicating that a substantial but tractable subset of all clinical features contributed to prediction. This sparsity is clinically meaningful: it identifies which conditions and medication categories are most strongly associated with hospitalisation risk in the elderly cohort, providing actionable insight for clinical audit and prioritisation.

%-----------------------------------------------

Table \ref{tab:model_performance_metrics} reports discrimination and calibration metrics for all three models on the held-out test set (N = 63,347), following Platt (log-loss) calibration. To address class imbalance, TG-CNN incorporated inverse-frequency class weighting into its training loss. For the primary held-out test set evaluation, LASSO and Random Forest were likewise fitted using inverse-frequency weighting to ensure balanced, directly comparable treatment across all three architectures. We note that applying isotonic regression as an alternative recalibration method reveals that TG-CNN's calibration curve becomes unstable and collapses for predicted probabilities above 0.5, a pattern not observed for LASSO or Random Forest, both of which remain well-behaved under isotonic recalibration. This instability indicates that TG-CNN's uncalibrated output distribution is poorly structured even for flexible, non-parametric recalibration, reinforcing LASSO's advantage in direct clinical deployability.
%Because TG-CNN requires temporal event sequences and cannot be trained on static-only inputs, static demographic performance is reported using a baseline neural network equivalent to the demographic branch of the full TG-CNN model (Table~\ref{tab:model_performance}).

%-------------------------------------

\subsection{Qualitative Analysis}

\begin{figure}
  \centering
  \includegraphics[width=0.99\textwidth]{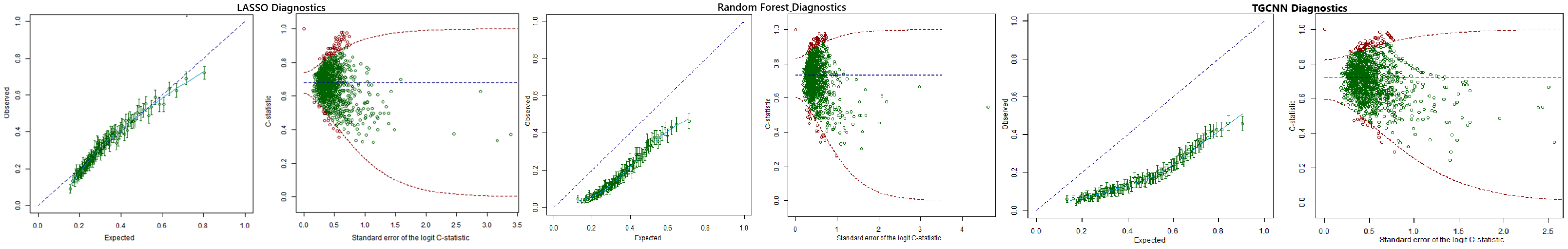}
  \caption{Diagnostics for (a) LASSO (b) Random Forest, and (c) TGCNN models with Calibration plots (on left) showing predicted versus observed hospitalisation rates; the curve falls below the ideal diagonal indicating systematic overestimation of risk. And GP-stratified C-statistic plots (right) showing performance across primary care practices; red points indicate practices outside the confidence envelope.}
  \label{fig:all_diagnostics}
\end{figure}

%\subsubsection*{LASSO Model Diagnostics}
\noindent\textbf{LASSO Model Diagnostics:} The calibration plot for the LASSO model (Figure~\ref{fig:all_diagnostics}) indicates that predicted probabilities correspond closely to observed event rates across the full risk spectrum, with the calibration curve tracking near the ideal diagonal. This is a
property of particular importance for clinical deployment where risk scores will be used to guide patient prioritisation for Structured Medication Reviews: a well-calibrated model means a patient assigned a 30\% risk score genuinely has approximately 30\% probability of hospitalisation, allowing clinicians to meaningfully interpret and act on individual predictions without further adjustment. The GP-stratified analysis shows that LASSO model performance is reasonably consistent across different primary care practices, with the majority of practices clustering around the overall mean performance line (blue dashed). A small number of practices at higher standard errors show wider variability, consistent with lower patient volumes at those sites. Overall, this supports the generalisability of the LASSO approach across heterogeneous primary care settings
within the CPRD population.

%\subsubsection*{Random Forest Model Diagnostics}
\noindent\textbf{Random Forest Model Diagnostics:}
The calibration plot for the Random Forest model (Figure~\ref{fig:all_diagnostics}) reveals a systematic pattern of miscalibration: predicted probabilities are consistently higher than observed event rates across all risk deciles. The calibration curve falls noticeably below the ideal diagonal, indicating that the model overestimates
hospitalisation risk, particularly at higher predicted probability values. While Random Forest achieves the best discrimination (AUC-ROC: 0.734), this calibration gap is an important limitation for clinical deployment: raw predicted probabilities from the Random Forest cannot be used directly as risk estimates without post-hoc recalibration, for example via Platt scaling or isotonic regression. This is reflected in Appendix \ref{appendix_A} (Table~\ref{tab:multicriteria})
where Random Forest is marked as not directly deployable without post-hoc correction, despite its leading AUC-ROC. The GP-stratified performance shows broadly similar consistency to the LASSO model across practices, with most
practices clustering around the overall mean C-statistic. However, a subset of practices show C-statistics below 0.5, suggesting the Random Forest may be less robust at practice level than its aggregate AUC-ROC implies (quantified in Appendix \ref{appendix_A}).

%However, the wider spread of practice-level performance estimates and the presence of several practices with C-statistics below 0.5 suggest that the Random Forest may be less robust at practice level than its aggregate AUC-ROC implies. This has direct implications for clinical deployment, where per-practice validity is as important as overall discrimination.

\begin{table} %[htbp]
\caption{Model performance comparison across  metrics on test set ($N_{\text{test}} = 63{,}347$).}
\label{tab:model_performance_metrics}
\centering
\begin{tabular}{lccccc}
\toprule
\textbf{Model} & \textbf{C-slope} & \textbf{Brier score} & \textbf{ECE} & \textbf{AUC-ROC} & \textbf{AUC-PR} \\
& \textbf{($\uparrow$1.0 best)} & \textbf{($\downarrow$ best)} & \textbf{($\downarrow$ best)} & \textbf{($\uparrow$ best)} & \textbf{($\uparrow$ best)} \\
\midrule
LASSO& \textbf{0.817} & \textbf{0.132} & 0.013 & \textbf{0.733} & \textbf{0.380} \\
Random Forest & 0.759 & 0.135 & 0.009 & 0.710 & 0.333 \\
TG-CNN & 0.391 & 0.137 & \textbf{0.008} & 0.702 & 0.334 \\
\bottomrule
\end{tabular}
\end{table}

%Random Forest model diagnostics. (a) Calibration plot showing predicted   versus observed hospitalisation rates; the curve falls below the ideal diagonal   indicating systematic overestimation of risk. (b) GP-stratified C-statistic plot   showing performance across primary care practices; red points indicate practices   outside the confidence envelope.
  
%\subsubsection*{TG-CNN Model Diagnostics}
\noindent\textbf{TG-CNN Model Diagnostics:} The calibration plot for TG-CNN (Figure \ref{fig:all_diagnostics}) shows the same qualitative pattern of miscalibration seen for Random Forest, predicted probabilities consistently exceed observed event rates across risk deciles, with the degree of overestimation at least as pronounced. This reinforces that systematic risk overestimation is not unique to one model class but a shared limitation of both higher-complexity approaches in this setting. GP-stratified performance shows a similarly wide spread of practice-level C-statistics to Random Forest, with neither model demonstrating superior consistency across primary care practices.

%The calibration plot for TG-CNN (Figure~\ref{fig:all_diagnostics}) reveals a pattern of miscalibration broadly similar to that of Random Forest: predicted probabilities are consistently higher than observed event rates across all risk deciles, with the calibration curve falling substantially below the ideal diagonal.
%The degree of overestimation appears at least as pronounced as that observed for Random Forest, reinforcing that this is not a property unique to one model class but a shared limitation of both higher-complexity approaches in this setting. The GP-stratified performance shows a similar distribution of practice-level C-statistics to the Random Forest, with the majority of practices clustering around the overall mean (blue dashed line). The spread of practice-level estimates and presence of outlying practices suggest comparable geographic variability to Random Forest, with neither model demonstrating superior consistency across primary care practices.

%%%%%%%%%%%%%%%%%%%%%%%%%%%%%%%%%%%%%%%%%%%%%%%%%%%%%%%%%%%%%%%%%%%%%%%%%%%
%%%%%%%%%%%%%%%%%%%%%%%%%%%%%%%%%%%%%%%%%%%%%%%%%%%%%%%%%%%%%%%%%%%%%%%%%%%

\section{Discussion}
We scope this negative result specifically to TG-CNN rather than to temporal deep learning architectures in general. TG-CNN was selected for its use of sparse linear algebra operations, which makes it computationally feasible on health-system-scale hardware unlike transformer-scale architectures such as BEHRT or RETAIN, despite TG-CNN demonstrating competitive performance in comparable EHR prediction settings \citet{hancox2024_review}).

%We scope this negative result specifically to TG-CNN rather than to temporal deep learning architectures in general. TG-CNN was selected because its use of sparse linear algebra operations makes it computationally feasible on health-system-scale hardware, unlike transformer-scale architectures such as BEHRT or RETAIN, which are not practically deployable in equivalent resource-constrained clinical computing environments, despite TG-CNN demonstrating competitive performance in comparable EHR prediction settings (detailed in existing review of literature by \citet{hancox2024_review}).

\subsection{Underperformance of Complex CNN methods}
Several factors explain, to some extent, the underperformance of TG-CNN relative to the simpler baselines: \textbf{Sparsity and high dimensionality of primary care EHR timelines}: Unlike dense clinical datasets such as ICU time-series, primary care records are characterised by sparse, irregularly sampled events over long observation windows, spread across hundreds of distinct clinical event and medication codes. The temporal weighting mechanisms of TG-CNN were originally designed for denser sequential datasets (e.g. online clickstream data \citep{hancox2022tgcnn}), and the exponential decay function may not be well-calibrated for event sequences spanning months or years. In this high-dimensional, noisy setting, the inductive biases of random forests (bagging and feature subsampling) and LASSO (explicit sparsity regularisation) may confer practical advantages over graph-based architectures whose learning dynamics are more sensitive to hyperparameter choices. \textbf{Prediction task vs. representation learning}: Previous research \citet{aslam2025clustering} demonstrated that TG-CNN produces superior trajectory clusters compared to Word2Vec and autoencoders on the ACT-MOOC dataset. However, producing well-separated latent clusters does not guarantee discriminative power for a specific binary label. The relationship between representation quality and downstream classification performance is non-trivial and task-dependent. \textbf{Training data constraints}: TG-CNN involves substantially more parameters than logistic regression and requires more data to learn effectively. Class imbalance, even when mitigated by inverse-frequency weighting, may disproportionately affect the convergence of deep architectures on binary outcomes with low base rates. \textbf{Static risk signal dominates over temporal ordering}: Hospitalisation risk over a 12-month horizon in an elderly, multimorbid cohort is likely driven predominantly by the accumulated burden of chronic conditions and polypharmacy patterns, a largely static, cross-sectional signal, rather than by the fine-grained temporal ordering of clinical events. If the predictive signal is largely ``what conditions and medications does this patient have'', rather than "in what sequence did they occur," temporal architectures such as TG-CNN offer no meaningful advantage over feature-based static baselines. We incorporated order-shuffled sequence ablations to isolate the specific contribution of temporal event sequencing (Appendix \ref{appendix_A}).

\subsection{Clinical Implications and Insights for ML in Healthcare}
From a clinical deployment perspective, the LASSO model offers substantial advantages beyond its competitive predictive performance. It is computationally lightweight, can be re-trained rapidly as new data accrue, and its non-zero coefficients provide a transparent explanation for any individual risk score. Healthcare professionals within the parent research programme have consistently expressed a preference for explainable AI that supports rather than replaces clinical reasoning, favouring transparent, well-calibrated models over marginally higher-discriminating black-box alternatives (\citet{abuzour2024plosone}; \citet{abuzour2026jmir}). A model whose outputs can be traced to specific conditions and medication categories, and communicated clearly in a patient history visualisation, is far more likely to gain clinical trust and adoption than a non-transparent graph neural network. This resonates with the parallel clustering work within the same research programme \citep{chapman2026clustering}, where LASSO Cox regression identified 74 interpretable patient features associated with ADR hospitalisation risk, enabling the derivation of five clinically meaningful patient subgroups. The interpretability of these features allowed clinicians to immediately map statistical findings onto clinical practice patterns, precisely the kind of actionable insight that the project seeks to provide. Our findings reinforce the importance of rigorously benchmarking complex models against well-tuned classical alternatives before concluding that deep learning offers a clinical benefit, particularly given the real infrastructure, maintenance, and governance costs of deploying unnecessarily complex models in resource-constrained healthcare systems \citet{rudin2019stop}. Our results suggest that for all-cause emergency hospitalisation prediction in primary care, this burden has not yet been met by the TG-CNN architecture in its current form. This motivate future work for understanding precisely when and why temporal architectures fail/succeed is scientifically valuable.

\subsection{Limitations and Future Directions}
Some of the limitations of this work include the following. First, TG-CNN results can be improved with a large number of GPUs, which is impractical for resource-constrained settings. While class weighting was uniformly applied across all final models evaluated on the held-out test set (Table \ref{tab:model_performance_metrics}), cross-validation exploratory runs (Table \ref{tab:model_performance}) relied on stratified cross-validation for traditional baselines. Future work will systematically evaluate how loss-weighting vs. unweighted training influences calibration curves and discrimination trajectories across all stages of cross-validation. Second, we evaluated only all-cause hospitalisation; for more specific outcomes where temporal disease progression patterns are more directly causal (e.g. specific ADR types, disease exacerbations), CNN may demonstrate a more meaningful advantage. Third, the cohort is restricted to patients aged 65+ in a specific CPRD observation window; generalisability to younger MLTC populations or different observation periods requires further study. Fourth, the CPRD Aurum data, while comprehensive, may not capture all relevant clinical events recorded in secondary care or community settings \citep{wolf2019cprd}. This study was conducted entirely within CPRD Aurum. While Aurum covers approximately half of all UK GP practices and provides substantial demographic and geographic heterogeneity, this does not constitute external validation, and future work should assess generalisability using an independent primary-care database such as SAIL or OpenSAFELY. We also did not conduct a subgroup or fairness analysis (e.g., stratified by ethnicity, deprivation, or age band); given known disparities in EHR data quality and coding practice across patient subgroups, this is an important direction for future work.

Building on the sparsity and static-signal limitations identified in Section 6.1, future work will investigate architectures specifically adapted to sparse, long-horizon primary care sequences, including hybrid approaches combining interpretable feature extraction (e.g. LASSO-selected features) with lightweight temporal encoders. We will also explore whether TG-CNN's strength in trajectory clustering can be leveraged for patient stratification upstream of risk prediction, enabling group-specific models that outperform population-level classifiers. Engagement with clinicians through the co-design process will continue to shape the design of AI tools so that interpretability and clinical usability remain central alongside predictive performance.

\begin{ack}
This work is funded by DynAIRx project. DynAIRx has been funded by the National Institute for Health and Care Research (NIHR) Artificial Intelligence for Multiple Long-Term Conditions (AIM) call (NIHR 203986). Additional support was provided by the University of Sheffield IJC Research Stimulation Fund. The views expressed in this publication are those of the author(s) and not necessarily those of the NIHR or the Department of Health and Social Care.
\end{ack}

% \section*{References}
% \bibliography{tgcnn_complexity_ref.bib}

\bibliographystyle{plainnat}
\bibliography{tgcnn_complexity_ref}

%%%%%%%%%%%%%%%%%%%%%%%%%%%%%%%%%%%%%%%%%%%%%%%%%%%%%%%%%%%%

\appendix
\section{Appendix}
%\section{Appendix A}
\label{appendix_A}

\section*{TG-CNN Architecture, Input Definitions, and Training Diagnostics}
\begin{figure}[h]
  \centering
  \includegraphics[width=0.99\textwidth]{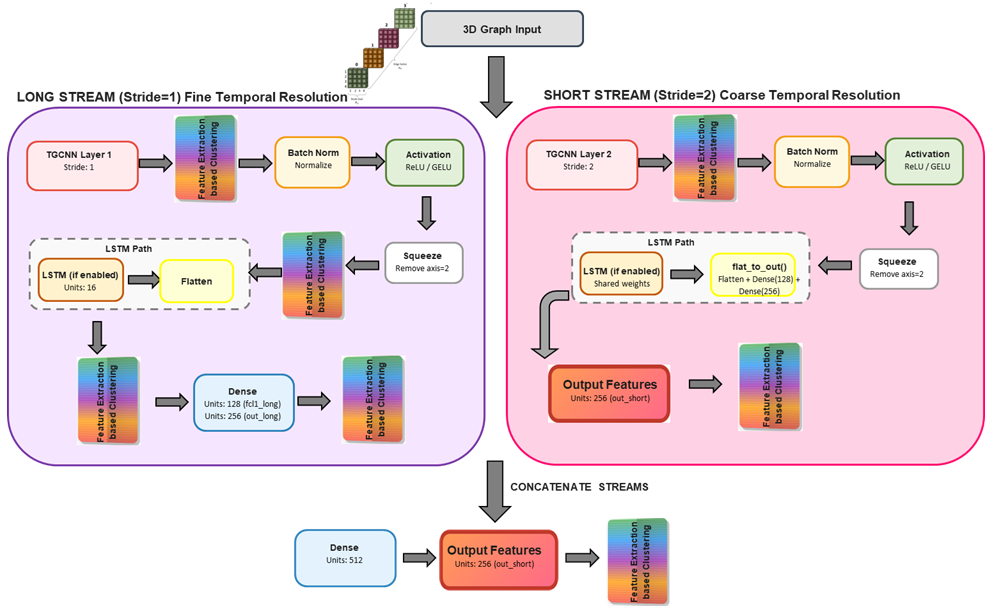}
  \caption{TG-CNN Architecture with Dual Layer based Configuration flexibility}
  \label{fig:config_tgcnn}
\end{figure}

TGCNN architecture designed used for this work is shown in Figure \ref{fig:config_tgcnn}. Please find definitions that are used in this paper as follows:

\noindent\textbf{Events:}
Binary features for prescriptions (BNF codes, n = 802) or condition codes (as per DynAIRx codelist; n = 190 unique codes). For LASSO regression and Random Forests, only binary feature labels were used to train models, regardless of when the event occurred. For TGCNN, dates associated with codes were also encoded (as described in Methods) and used in model training.

\noindent\textbf{Demographics:}
19 features to describe patient demographics, including age (linearly scaled using the following formula to ensure similar scale for all features: (age-65)/25 ), Index of Multiple Deprivation (IMD; one-hot encoded categories including one for missing values), ethnicity (one-hot encoded categories) and sex.

\noindent\textbf{Events \& Demographics}:
Both data types above used for model development.

%The final feature space comprises clinical condition codes observed in the elderly cohort (drawn from the broader DynAIRx framework based on cohort prevalence \cite{aslam2025automation}), British National Formulary (BNF) prescription codes, and demographic features. Table~\ref{tab:feature_inventory} consolidates these input categories and is referenced from both the Methods and the Appendix.

\begin{figure} %[h]
  \centering
  \includegraphics[width=0.99\textwidth]{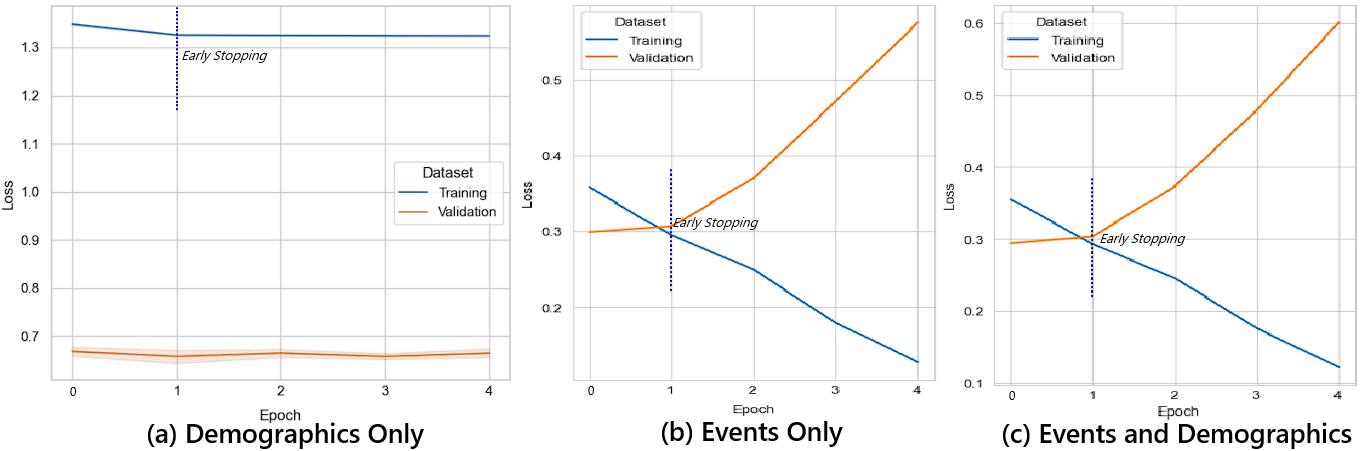}
  \caption{TGCNN Training for Demographics, Events, and Events with Demographics. Dashed vertical lines indicate the early stopping epoch (Epoch 1) where validation loss reached its minimum and model weights were saved for testing}
  \label{fig:all_tgcnn_combined}
\end{figure}

Training graphs for TGCNN are shown in Figure \ref{fig:all_tgcnn_combined}. The reported results correspond to model weights at the epoch of minimum cross-validation loss (epoch 1). Training-curve figures in the Appendix are annotated to indicate this early-stopping point.

Table~\ref{tab:multicriteria} summarises all three evaluation criteria across models: LASSO is the only model that is both well-calibrated and directly deployable without post-hoc correction.

\section*{Feature Inventory}
The final feature space comprises clinical condition codes observed in the elderly cohort (drawn from the previous DynAIRx framework based on cohort prevalence \cite{aslam2025automation}), British National Formulary (BNF) prescription codes, and demographic features, as summarized in Table~\ref{tab:feature_inventory}. The original condition list comprised 260 conditions. During codelist construction, drug-based related conditions were merged along with clinicians feedback; the DynAIRx repository consist of non-overlapping 211 codelists. Of these 211, 12 had zero observed occurrences in the elderly cohort and were excluded: Alopecia, Alopecia areata, ankylosing spondylitis, body mass index, chronic dermatitis/eczema, primary pulmonary hypertension, tuberculosis, skin ulcer, renal and bladder stones, occupational lung diseases, liver disease (unknown), and hypotension/syncope. Many of these would be expected in this cohort given related prescribing evidence, for example, 1,663 patients had a prescription code for anti-tuberculosis medication (BNF 5.1.9) despite zero tuberculosis diagnosis codes being observed suggesting a systematic encoding issue in the diagnosis-code data, though the specific mechanism could not be established and is flagged here as a limitation warranting further investigation. A further 9 codelists were condensed into combined labels in this study to reduce redundancy: five cholesterol-related codes were combined into a single 'cholesterol' feature; 'history and monitoring' codes for haemorrhagic, ischaemic, subarachnoid haemorrhage, and transient ischaemic attack stroke types were condensed with their corresponding disease labels; and 'ischaemic heart disease' and 'ischaemic heart disease history' were condensed. This yields the final set of 190 condition codes summarised in the feature inventory.

\begin{table}
\centering
\caption{Multi-criteria Model Summary.}
\label{tab:multicriteria}
\small
\begin{tabularx}{\textwidth}{>{\raggedright\arraybackslash}p{3.5cm}|>{\centering\arraybackslash}p{2cm}|>{\raggedright\arraybackslash}X|>{\centering\arraybackslash}p{2cm}|>{\raggedright\arraybackslash}p{3cm}}
\toprule
\textbf{Model} & \textbf{Best AUC-ROC} & \textbf{Calibration} & \textbf{Interpretable} & \textbf{Deployable without post-hoc correction?} \\
\midrule
LASSO & 0.733 & Well-calibrated & Yes (548 coefficients) & Yes \\
Random Forest & 0.710 & Miscalibrated & Partial & No \\
TG-CNN & 0.702 & Miscalibrated  & No (black box) & No \\
\bottomrule
\end{tabularx}
\end{table}

\begin{table} %[htbp]
\centering
\caption{Consolidated Feature Inventory Across Model Input Categories}
\label{tab:feature_inventory}
\begin{tabular}{lcm{7cm}}
\hline
\textbf{Feature Category} & \textbf{Count} & \textbf{Description / Encoding} \\ \hline
Clinical Conditions & 190 & Unique condition codes observed in the elderly cohort (DynAIRx framework) \\
Medication Prescriptions & 802 & Unique British National Formulary (BNF) prescription codes \\
Demographics & 19 & 18 binary encodings plus continuous scaled age \\ \hline
\end{tabular}
\end{table}

\section*{Test-Set Discrimination and Calibration}
\begin{figure} %[h]
  \centering
  \includegraphics[width=1\textwidth]{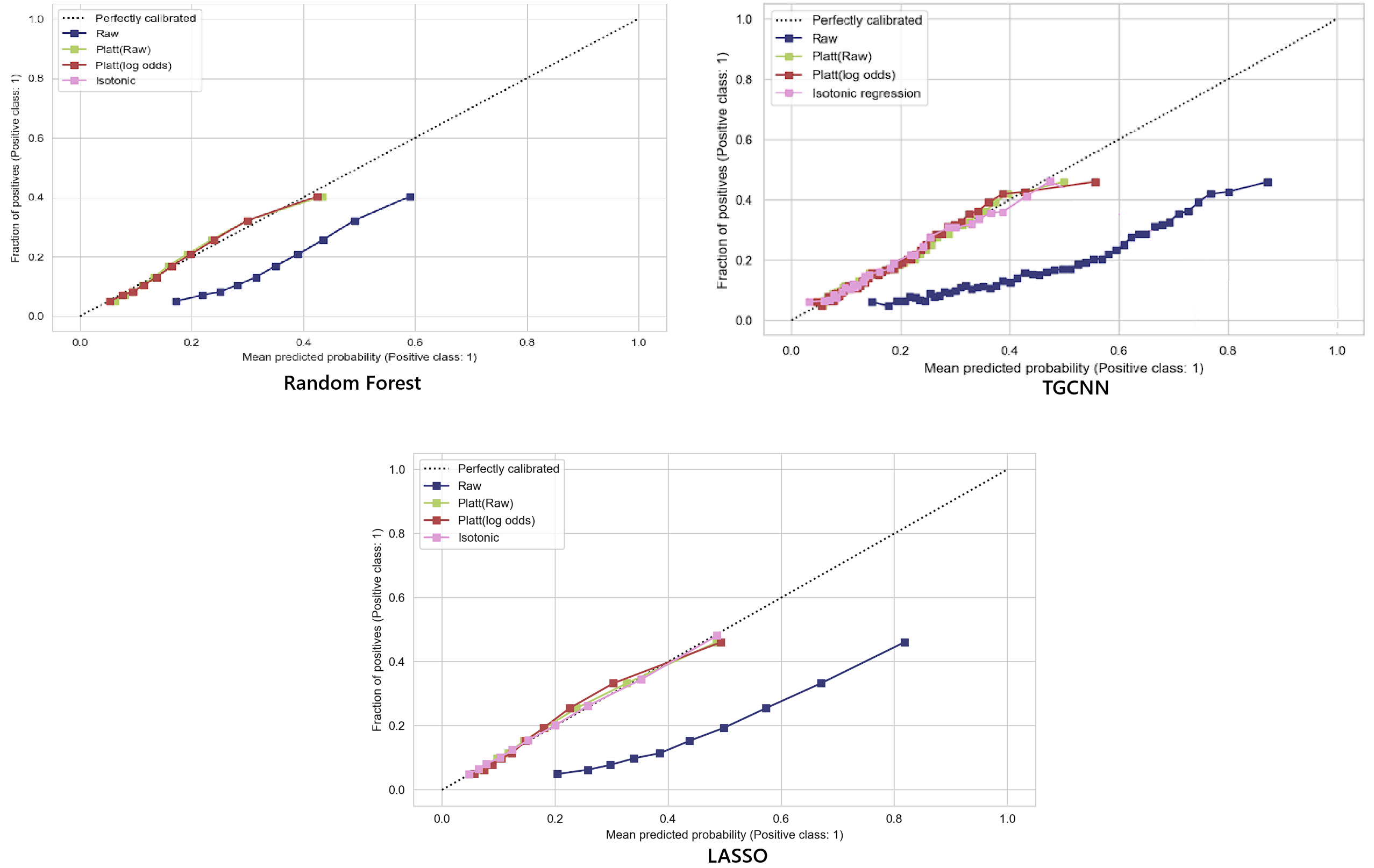}
  \caption{Calibration curves for Logistic Regression (LASSO), Random Forest, and TG-CNN models on the test set, comparing raw predictions against post-hoc Platt scaling and Isotonic regression.}
  \label{fig:calibration_curves}
\end{figure}

Model performance and calibration results were evaluated across the unseen test partition ($n \approx 63,000$), with calibration curves for Logistic Regression (LASSO), Random Forest, and TG-CNN displayed in Figure \ref{fig:calibration_curves}. To prevent data leakage and ensure uncorrupted performance estimates, all calibration transformations—including Platt scaling on raw predictions, Platt scaling on log-odds, and non-parametric Isotonic regression were fitted on a dedicated, held-out 20\% subset of the training data. 

Uncalibrated raw models systematically overestimated risk across the probability spectrum. Post-hoc recalibration via Platt scaling and isotonic regression markedly improved agreement with observed outcome proportions, aligning predicted probabilities closely along the ideal 45-degree diagonal across all three model architectures. Specifically, both parametric (Platt scaling) and non-parametric (Isotonic regression) methods effectively re-aligned predicted probabilities with observed event frequencies up to predicted risks of $0.5$. For TG-CNN (Figure \ref{fig:calibration_curves}, top right), while Isotonic regression captured mid-range probabilities effectively, Platt scaling provided superior overall stability by avoiding the tail-end artifact caused by non-parametric step-fitting in sparse higher-probability bins.  

Overall, Logistic Regression (LASSO) emerged as the top-performing and most reliable model. It achieved the highest discrimination power ($\text{AUC-ROC} = 0.733$, $\text{AUC-PR} = 0.380$), the best calibration slope ($0.817$), a low Expected Calibration Error ($\text{ECE}$), and the lowest overall prediction error ($\text{Brier score} = 0.132$). Furthermore, following post-hoc Platt scaling, Logistic Regression demonstrated excellent calibration stability across the entire risk spectrum without suffering from high-probability tail artifacts. Consequently, Logistic Regression (LASSO) calibrated via Platt scaling represents the most robust and clinically actionable model in this benchmark.

%\section*{Practice-Level (GP) Discrimination}
\noindent\textit{Practice-Level (GP) Discrimination}: Across GP practices, 42.1\% had a Random Forest C-statistic below 0.5, indicating discrimination no better than chance at those sites. Practices with C-statistic below 0.5 had smaller patient lists (mean $39.7 \pm 33.4$ patients) than other practices (mean $45.7 \pm 32.9$ patients), and lower hospitalisation outcome prevalence ($16.1\% \pm 9.4\%$ vs. $20.0\% \pm 7.7\%$). Both smaller sample size and lower event prevalence increase the variance of any discrimination estimate, and this pattern is consistent with reduced statistical power at low-volume, low-event practices rather than a distinct clinical or coding-related failure mode specific to those sites. We did not identify a systematic difference in coding patterns between the two groups; this remains a direction for further investigation.

%Held-out test set calibration curves across models. Risk calibration plots for Random Forest (top left), TG-CNN (top right), and LASSO (bottom) comparing uncalibrated predictions (Raw) against post-hoc recalibration via Platt (Raw), Platt (log odds), and Isotonic regression against the line of perfect calibration (dotted black)

\section*{Order-Shuffled TG-CNN Ablation}
\begin{figure} %[h]
  \centering
  \includegraphics[width=1\textwidth]{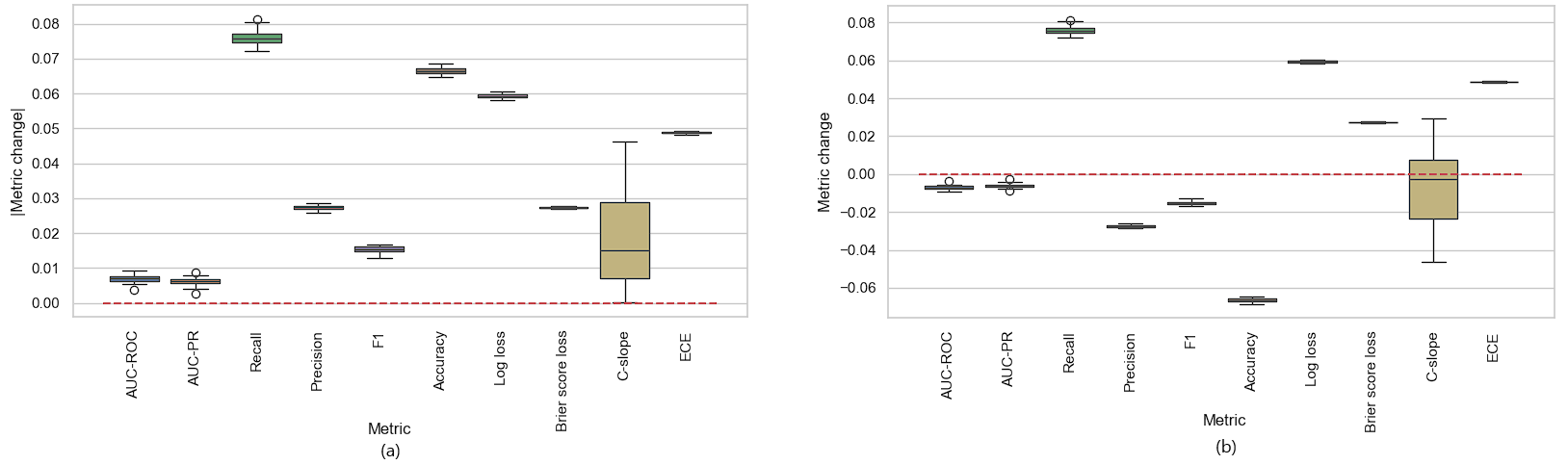}
  \caption{Change in TG-CNN test-set performance and calibration metrics when event order is shuffled, relative to the unshuffled baseline (dashed red line at zero), across 50 independent shuffles. (a) Absolute value of the change in each metric. (b) Raw change in each metric, showing whether shuffling increased or decreased each value.}
  \label{fig:shuffling}
\end{figure}

To isolate the specific contribution of temporal event sequencing, we conducted an order-shuffled ablation on the held-out test set ($n \approx 63,000 $). Across 50 independent shuffles, the order of events within each patient's sequence was randomised while preserving the same set of event codes; TG-CNN inference was re-run on each shuffled test set using the cross-validation-selected model weights, and evaluation metrics were recomputed against the unshuffled (original-order) baseline (Figure \ref{fig:shuffling}). Shuffling event order degraded AUC-ROC, AUC-PR, precision, F1, and accuracy, and worsened log loss, Brier score, and Expected Calibration Error (ECE), relative to the unshuffled baseline. This isolates a clear, measurable contribution from event ordering: TG-CNN does not treat its input as an unordered set of codes, and disrupting the sequence structure meaningfully harms performance and calibration. This contribution is nonetheless bounded. TG-CNN's unshuffled performance (AUC-ROC 0.702, Table \ref{tab:model_performance_metrics}) still trailed both LASSO (0.733) and Random Forest (0.710) on the same test set. Temporal sequencing therefore improves TG-CNN over its own shuffled baseline without closing the gap to the simpler, static baselines, event ordering is informative, but not sufficient to make TG-CNN the best-performing model in this comparison.

Recall increased under shuffling i.e. the unshuffled (intact-order) model shows lower recall, and more false negatives, than the shuffled version. This suggests TG-CNN's temporal weighting may systematically under-weight clinically informative events occurring earlier in a patient's history relative to more recent ones. UK primary-care coding conventions often fully code a condition at first diagnosis but do not necessarily re-code it at every subsequent visit, since repeat documentation of an already-known, stable condition is frequently treated as redundant, a pattern especially pronounced in elderly, multimorbid patients, where not every active condition is re-coded at each visit. Combined with TG-CNN's recency-weighted decay ($G(i,j,k) = \exp(-\gamma t)$), this means an important condition recorded only once, early in a patient's history, may be systematically down-weighted relative to more recent, but less clinically significant, records. This offers a plausible explanation for the higher false-negative rate observed under the model's default, unshuffled temporal weighting, compared to the shuffled condition. Residual connections, which help preserve information from earlier layers or timesteps in deep architectures, and fusion of temporal models with binary event representations, of the kind used by LASSO and Random Forest, are promising directions for future work to address this limitation.

%Combined with TG-CNN's recency-weighted decay ($G(i,j,k) = \exp(-\gamma t)$), this could leave earlier but clinically important events under-weighted relative to more recent records, consistent with the increased false-negative rate observed under the model's default (unshuffled) temporal weighting. 

%Feature importance was evaluated using permutation importance, defined as the mean increase in loss when a feature's values are randomly shuffled on the held-out test set (Figure \ref{fig:permutations}). Given TG-CNN's comparatively poor calibration and unsuitability for direct clinical deployment, we focus our primary interpretability analysis on LASSO our recommended deployable model using both permutation importance and SHAP values, comparing against Random Forest baseline patterns.  Across both models, primary care medication-related BNF codes (e.g., event\_bnf\_020802, event\_bnf\_090604) and functional frailty markers specifically recorded indicators of reduced mobility (event\_Housebound) dominate the top-ranked predictors. While established chronic conditions such as event\_Dementia and event\_Atrial Fibrillation contribute to risk estimates, individual diagnosis and condition codes do not rank among the primary top predictors. This demonstrates that the predictive signal stems overwhelmingly from cumulative medication burden and functional status rather than any single disease diagnosis. These findings align closely with prior hospitalisation risk modeling in similar primary care cohorts \cite{fahmi2023combinations}.

\section*{Interpretability: Permutation Importance and SHAP Analysis}
\begin{figure}
  \centering
  \includegraphics[width=1\textwidth]{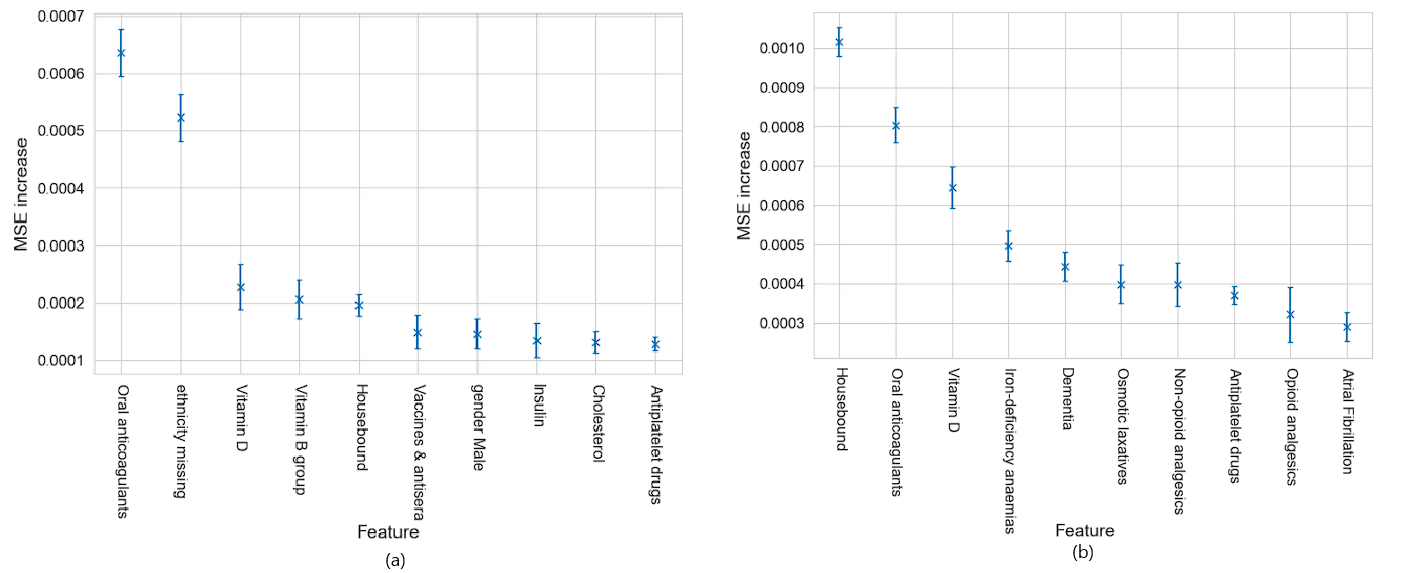}
  \caption{Permutation feature importance for LASSO (top predictors: primary care prescription codes, BNF, and functional status markers, event\_Housebound) and Random Forest (which additionally ranks individual chronic conditions, including dementia and atrial fibrillation, among its top predictors), ranked by mean MSE increase upon feature permutation on the held-out test set.}
  \label{fig:permutations}
\end{figure}

\begin{figure}
  \centering
  \includegraphics[width=1\textwidth]{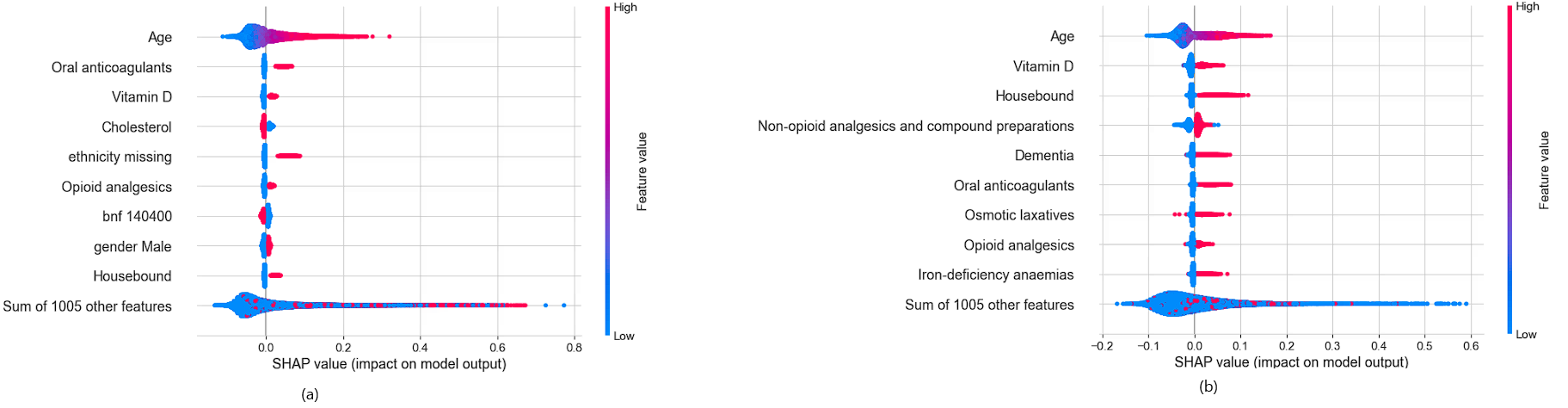}
  \caption{SHAP values for LASSO and Random Forest on the held-out test set, showing age as the dominant predictor for both models, followed by medication-related and functional status features consistent with the permutation importance results.}
  \label{fig:shap}
\end{figure}

Feature importance was evaluated using permutation importance, defined as the mean increase in loss when a feature's values are randomly shuffled on the held-out test set, and using SHAP values, on the held-out test set (Figure \ref{fig:permutations}, Figure \ref{fig:shap}). We focus our primary interpretability analysis on LASSO, our recommended deployable model, comparing patterns against Random Forest.

For LASSO, permutation importance is dominated by medication-related BNF codes (oral anticoagulants, vitamin D, vitamin B group, insulin, cholesterol, antiplatelet drugs), missing ethnicity data, and housebound status; no individual diagnosis or condition code ranks among its top 10 predictors. Random Forest shows a similar reliance on medication and functional markers, housebound status and oral anticoagulants rank first and second for both models, but, unlike LASSO, also ranks two individual condition codes, dementia and atrial fibrillation, among its top ten predictors, alongside iron-deficiency anaemia. This suggests that while medication burden and functional status are the dominant predictive signal across both models, Random Forest additionally captures some diagnosis-specific risk that LASSO's linear structure does not surface as prominently.

%Model permutation feature importance. Top predictor rankings ordered by mean MSE increase upon feature permutation, highlighting the dominant contribution of primary care prescription sequences (BNF codes), functional status markers (event\_Housebound), and chronic comorbidities.

SHAP analysis identifies age as, by a wide margin, the most influential predictor for both models, followed by a similar cluster of medication and functional features (oral anticoagulants, vitamin D, housebound status) seen in the permutation importance results. These findings are broadly consistent with prior hospitalisation risk modelling in similar primary care cohorts \cite{fahmi2023combinations}.

\begin{table*} %[t]
\centering
\begin{threeparttable}

\caption{Hyperparameter search space with selected hyperparameters for the final model.}
\label{tab:hyperparameters}
\small
\begin{tabular}{p{2.7cm} p{4.7cm} p{3.8cm} p{2.5cm}}
\toprule
\textbf{Category} & \textbf{Parameter} & \textbf{Values tested} & \textbf{Best model} \\
\midrule

\multirow{7}{*}{Learning process}
& Max epochs & 10 & 1\textcolor{red}{*}\\
& Initial learning rate & $\{10^{-2},\,10^{-3}\}$ & $10^{-3}$ \\
& \multirow{2}{*}{Learning rate decay} & Exponential ($\gamma=0.9$& \multirow{2}{*}{Same} \\
& & decay every $10^6$ steps) & \\
& Optimizer & Adam & Adam \\
& Batch size & 64 & 64 \\
& Activation function & \{GELU, ReLU\} & ReLU \\
& Class-weighted loss & \{True, False\} & True \\

\midrule

\multirow{2}{*}{Regularization}
& Dropout & \{0.5, 0.6, 0.7, 0.8, 0.9\}& 0.9\\
& L1/L2 regularization& \multirow{2}{*}{$\{10^{-2}, 10^{-3}, 10^{-4}\}$ }& \multirow{2}{*}{$10^{-4}$ }\\
& (ElasticNet-like) & & \\
\midrule

\multirow{5}{*}{TGCNN layer}
& Number of filters & \{4, 8, 16, 32\} & 8 \\
& Filter size & \{3, 4, 6\} & 4 \\
& Fully connected layer size & \{32, 64, 128\} & 32 \\
& Number of linear layers & \{1, 2\} & 1 \\
& Residual connections & \{True, False\} & True \\

\midrule

\multirow{2}{*}{LSTM}
& LSTM units & \{8, 12\} & 8 \\
& Shared LSTM (shared & \multirow{2}{*}{\{True, False\}} & \multirow{2}{*}{False} \\
& across stride-1/2 branches) &  & \\
\midrule

\multirow{3}{*}{Demographics}
& Demographics used & \{True, False\} & True \\
& Number of demographic layers & \{1, 2\} & 1 \\
& Units per demographic layer& \{32, 64, 128, 256, 512\} & 512 \\

\midrule

Final classifier & Use outputs of all FC& \multirow{2}{*}{\{True, False\}} & \multirow{2}{*}{True} \\
& layers in classifier &  & \\
\bottomrule

\end{tabular}
\begin{tablenotes}
\item[\textcolor{red}{*}] TGCNN converged immediately.
\end{tablenotes}

\end{threeparttable}

\end{table*}

% Table X reports the search space and final selected configuration for all TG-CNN hyperparameters. 

\section*{TG-CNN Hyperparameter Search and Final Configuration}
TG-CNN was tuned using the same 5-fold cross-validation protocol applied to LASSO and Random Forest, resolving the asymmetric tuning procedure used at initial submission. Hyperparameters were selected through a systematic search over the candidate values summarized in Table~\ref{tab:hyperparameters}. Under this matched protocol, TG-CNN achieves a mean cross-validated AUC-ROC of 0.712 (SD 0.0022) with demographic features included, and 0.685 (SD 0.0029) without. Training employed the Adam optimizer with an initial learning rate of $10^{-3}$, a batch size of 64, and an exponential learning rate decay schedule with a decay factor of 0.9 applied every $10^6$ optimization steps (where each step corresponds to a mini-batch update). Although a maximum of 10 epochs was permitted, the TGCNN consistently converged within a single epoch. ReLU was selected over GELU as the activation function, and class-weighted cross-entropy loss was used to mitigate class imbalance by assigning greater importance to underrepresented classes. Model generalization was further improved through dropout and ElasticNet-style regularization, which combines L1 and L2 penalties to simultaneously encourage sparse and well-conditioned model weights.

For the network architecture, the search optimized the TGCNN filter configuration, the size and depth of the fully connected layers, and the temporal modeling strategy. The final model employed eight graph convolution filters of size four, followed by a single 32-unit fully connected layer with residual connections, where the layer input is added to its output to preserve the residual signal and facilitate optimization. Temporal dependencies were modeled using an 8-unit LSTM, with independent LSTMs for the stride-1 and stride-2 TGCNN branches rather than sharing weights between them. Demographic information was incorporated through a dedicated single-layer network with 512 hidden units. Finally, instead of relying solely on the final fully connected representation (i.e., a linear classifier), the outputs of all fully connected blocks were concatenated and provided to the final classifier, enabling predictions to leverage features learned at multiple levels of abstraction.

% \newpage
% \appendix
% \section*{Appendix A.}

% Some more details about those methods, so we can actually reproduce
% them.  After the blind review period, you could link to a repository
% for the code also.  \emph{MLHC values both rigorous evaluation as well
%   as reproduciblity.}

%%%%%%%%%%%%%%%%%%%%%%%%%%%%%%%%%%%%%%%%%%%%%%%%%%%%%%%%%%%%

% \newpage
% \input{checklist.tex}

\end{document}